\documentclass{article} % For LaTeX2e
\usepackage{colm2024_conference}

\usepackage{microtype}
\usepackage{hyperref}
\usepackage{url}
\usepackage{booktabs}
\usepackage{graphicx}%
\usepackage{multirow}%
\usepackage{amsmath,amssymb,amsfonts}%
\usepackage{amsthm}%
\usepackage{mathrsfs}%
\usepackage[title]{appendix}%
\usepackage{xcolor}%
\usepackage{textcomp}%
\usepackage{manyfoot}%
\usepackage{booktabs}%
\usepackage{algorithm}%
\usepackage{algorithmicx}%
\usepackage{algpseudocode}%
\usepackage{listings}%

\title{Automatic Interactive Evaluation for Large Language Models with State Aware Patient Simulator}

\colmfinalcopy

\author{Yusheng Liao, Yutong Meng, Yuhao Wang, Hongcheng Liu \\
Shanghai Jiao Tong University\\
\texttt{\{liao20160907, colane, hongcheng\_liu\}@sjtu.edu.cn} \\
\And
Yu Wang\thanks{Corresponding Author}, Yanfeng Wang \\
Shanghai Jiao Tong University \\
Shanghai Artificial Intelligence Laboratory \\
\texttt{\{wangyanfeng622, yuwangsjtu\}@sjtu.edu.cn} \\
}

\begin{document}

\maketitle

\begin{abstract}
Large Language Models (LLMs) have demonstrated remarkable proficiency in human interactions, yet their application within the medical field remains insufficiently explored. Previous works mainly focus on the performance of medical knowledge with examinations, which is far from the realistic scenarios, falling short in assessing the abilities of LLMs on clinical tasks. In the quest to enhance the application of Large Language Models (LLMs) in healthcare, this paper introduces the Automated Interactive Evaluation (AIE) framework and the State-Aware Patient Simulator (SAPS), targeting the gap between traditional LLM evaluations and the nuanced demands of clinical practice. Unlike prior methods that rely on static medical knowledge assessments, AIE and SAPS provide a dynamic, realistic platform for assessing LLMs through multi-turn doctor-patient simulations. This approach offers a closer approximation to real clinical scenarios and allows for a detailed analysis of LLM behaviors in response to complex patient interactions. Our extensive experimental validation demonstrates the effectiveness of the AIE framework, with outcomes that align well with human evaluations, underscoring its potential to revolutionize medical LLM testing for improved healthcare delivery.
\end{abstract}

\section{Introduction}
Large Language Models (LLMs) such as ChatGPT~\cite{elmohamed} and GPT-4~\cite{DBLP:journals/corr/abs-2303-08774} have significantly advanced many fields with their capability in understanding and generating natural language~\cite{kaddour2023challenges,hadi2023survey}. Despite these advancements, their impact in the medical domain has not been equally profound~\cite{tian2024opportunities,thirunavukarasu2023large,yang2023large}. While there have been targeted efforts to develop LLMs specifically for the medical field—such as Medalpaca~\cite{han2023medalpaca}, Huatuo~\cite{wang2023huatuo}, ChatDoctor~\cite{yunxiang2023chatdoctor}, and DoctorGLM~\cite{xiong2023doctorglm}—widespread practical deployment in clinical environments has not been achieved, with the notable exception of Med-PaLM 2~\cite{singhal2023large}. This gap primarily arises from a misalignment between the prevalent evaluation methods and the unique requirements of clinical contexts~\cite{zhou2023survey,minssen2023challenges,tu2024towards}, which obstructs the effective utilization of medical LLMs in healthcare consultations. Notably, existing benchmarks predominantly evaluate medical LLMs based on their knowledge performance via standardized examinations~\cite{DBLP:conf/emnlp/JinDLCL19,DBLP:conf/chil/PalUS22,DBLP:journals/corr/abs-2009-13081}, and do not adequately measure their proficiency in essential clinical tasks, such as pre-consultation and diagnostic support~\cite{yang2023large}. Moreover, the benchmarks designed for more complex clinical tasks often restrict these models to limited action spaces~\cite{wei-etal-2018-task,10.1093/bioinformatics/btac817} or reduce dynamic interactions to static question-answering scenarios~\cite{zhao2022medical, liu2022meddg, bao2023disc, zhang2023huatuogpt}, failing to capture the full complexity of real-world clinical interactions.~\cite{webster2023six}

In response to these challenges, some research initiatives have employed numerous human participants to role-play as patients in real-time interactions with LLMs, thus simulating clinical scenarios to evaluate the capabilities of these models~\cite{tu2024towards}. However, this approach incurs high costs and lacks scalability, limiting its utility for the extensive development and testing of medical LLMs~\cite{mehandru2024evaluating}. Therefore, there is a pressing need to develop an affordable yet effective methodology for automatically assessing the clinical competencies of LLMs in the healthcare domain.

\begin{figure*}[!t]
%是可选项 h表示的是here在这里插入，t表示的是在页面的顶部插入
\centering
\includegraphics[width=1.0\textwidth]{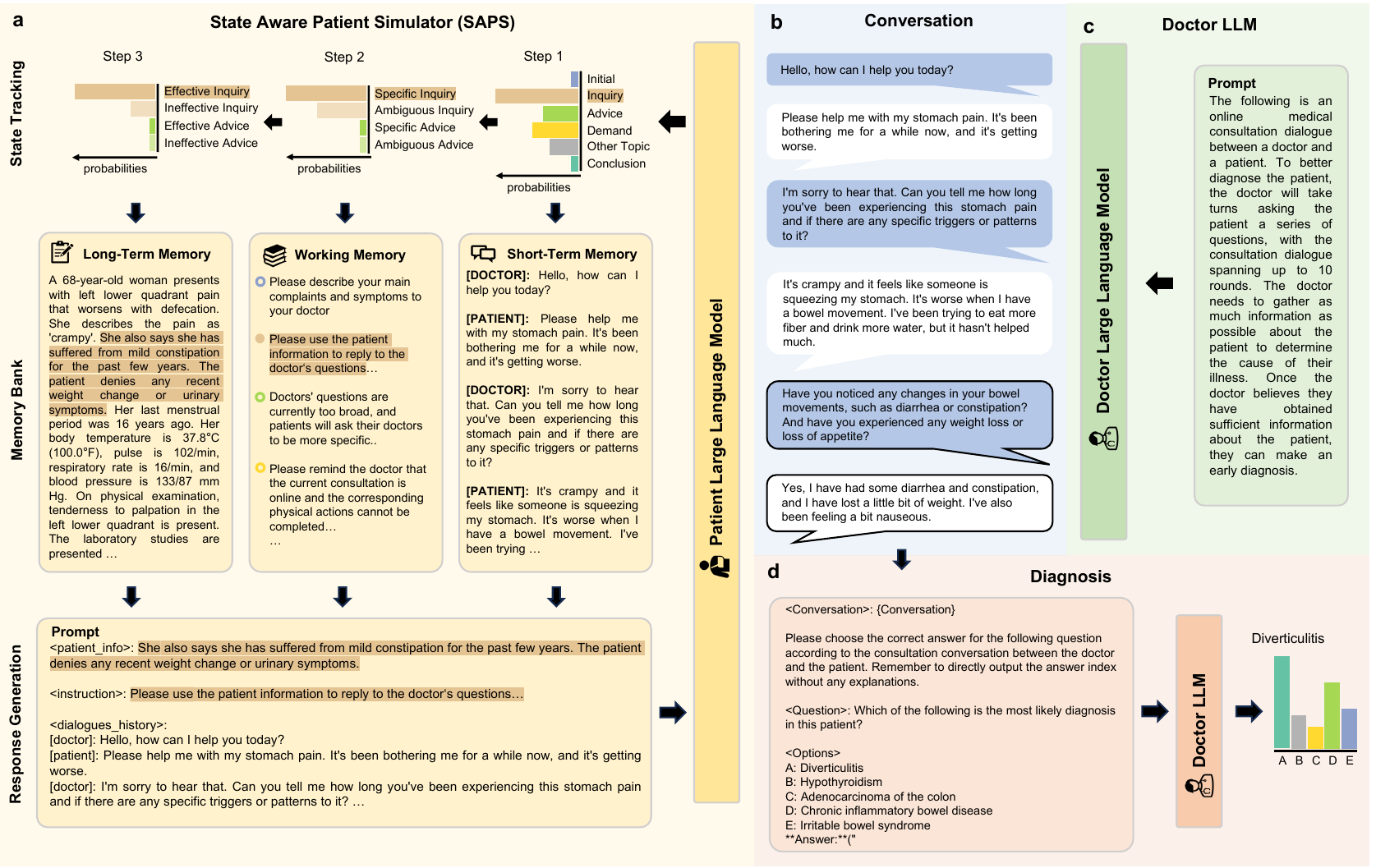}

\caption{\textbf{Overview of the Automatic Interactive Evaluation framework.} \textbf{a} State Aware Patient Simulator~(SAPS). SAPS structure includes a state tracker for classifying doctor behaviors, a memory bank for information retrieval, and a response generator for creating replies. Sentences with a dark background represent the parts that are activated within SAPS. \textbf{b} Conversation history between SAPS and Doctor LLM. The dialogue with a black border represents the latest round of dialogue. \textbf{d} Evaluated doctor LLM and its prompts. \textbf{e} Diagnosis. After the consultation dialogue, the doctor model must diagnose based on the information gathered during the conversation. `{Conversation}' indicates the dialogue history.} 
\label{fig: overview}
\end{figure*}

To address these gaps, in this paper, we propose a novel evaluation approach, Automatic Interactive Evaluation (AIE), as detailed in Figure~\ref{fig: overview}. This methodology leverages the role-playing capabilities of LLMs to function as patient simulators~\cite{shanahan2023role}, facilitating dynamic multi-turn interactions with the doctor LLMs being assessed. Within the AIE framework, the concept of `doctor LLMs' refers to large language models tested for their diagnostic interaction capabilities. These models must effectively gather comprehensive patient information through dialogue, culminating in a diagnosis or treatment recommendation. AIE aims to more faithfully simulate clinical scenarios, presenting a complex challenge that surpasses previous methodologies and offers a deeper assessment of LLM capabilities within a practical healthcare context. To our knowledge, this is the first evaluation that interactively validates the consultation abilities of LLMs through such a patient simulator.

Additionally, inspired by task-oriented dialogue systems~\cite{wen2016network}, we have developed a unique construct for the patient simulator named the State Aware Patient Simulator (SAPS). This  innovative simulator comprises three main components: a state tracker, a memory bank, and a response generator. Together, these components enable the realistic simulation of patient-doctor interactions. At each dialogue turn, the state tracker classifies each action taken by the doctor LLM into predefined categories, which allows SAPS to retrieve the most relevant information from the memory bank and generate appropriate responses. This structured approach enables SAPS to adapt its responses based on the varying behaviors of the doctor, facilitating a nuanced analysis of the decision-making processes typical in medical consultations. Importantly, we have defined 10 categories of actions, refined through preliminary experiments, based on their nature and effectiveness in clinical interactions. This categorization is critical for tailoring the responses of SAPS, ensuring that the simulated dialogues closely replicate authentic clinical conversations.

The empirical evaluation of AIE involves a two-step process. Initially, the effectiveness of SAPS is validated against a simulated patient test set derived from actual hospital cases, demonstrating its superior ability to mimic human patient behaviors accurately. Subsequently, the evaluation metrics, developed based on human and patient preferences and aligned with standards such as the UK General Medical Council Patient Questionnaire (GMCPQ) and principles from the consensus on patient-centered communication (PCCBP)~\cite{king2013best}, are assessed for their correlation with human judgments. This dual-phase evaluation underscores the potential of AIE and SAPS to revolutionize the assessment of medical LLMs, offering insights into their applicability in real-world clinical settings.

\begin{figure*}[!th]
%是可选项 h表示的是here在这里插入，t表示的是在页面的顶部插入
\centering
\includegraphics[width=1.0\textwidth]{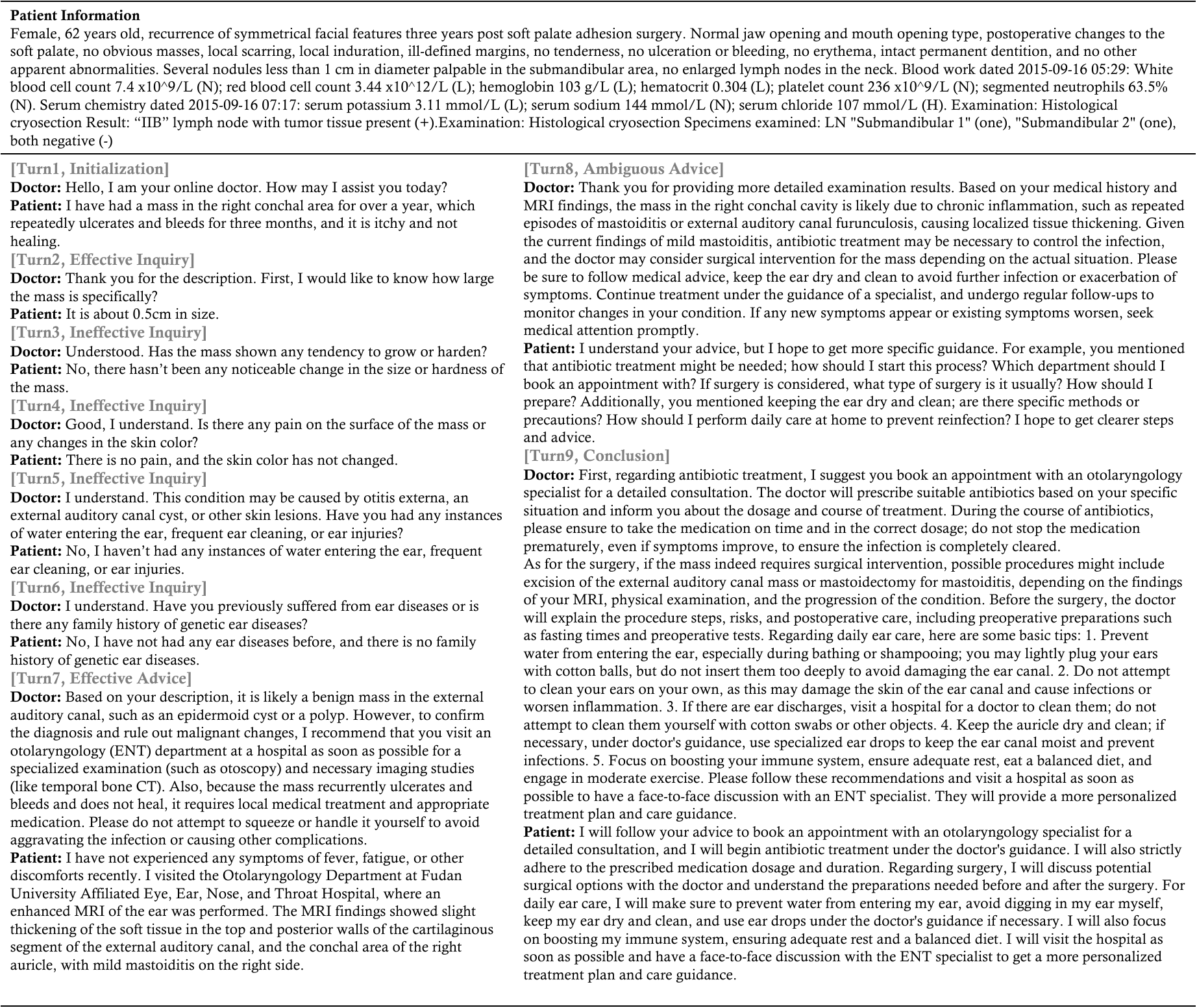}

\caption{Case examples of the consultation conversation between the doctor LLM Qianwen (noted as \textbf{Doctor}) and the patient simulator SAPS (noted as \textbf{Patient}). Each round is numbered and noted with the corresponding doctor LLM action category. The patient information is included above the conversation example.}
\label{fig: case_study}
\end{figure*}

\begin{table}[th]
\centering
% \resizebox{\textwidth}{!}{%
\caption{Definitions of the doctor LLM's action and the corresponding response requirements of the patient simulator. }
\label{tab:action and response definition}
\resizebox{0.95\textwidth}{!}{%
\begin{tabular}{ccc}
\toprule
\textbf{Doctor Actions} & \textbf{Actions Description}                                         & \textbf{Patient Required Behaviors}        \\
\midrule
Initialization         & The first action in the consultation conversation                                                    & Reply with chief complaint                 \\
Conclusion             & Make a diagnosis or reach the maximum conversation length                                            & None                                       \\
Effective Inquiry      & The inquired information is contained in the patient's information                                   & Reply with corresponding information       \\
Ineffective Inquiry    & The inquired information is not contained in the patient's information                               & Negation                                   \\
Ambiguous Inquiry      & The inquiry is too vague or too broad                                                                & Request the inquiry more specific          \\
Effective Advice       & The results of the recommended exams and test are contained in the patient's information             & Reply corresponding results                \\
Ineffective Advice     & The results of the recommended exams and test results are not contained in the patient's information & Follow advice                              \\
Ambiguous Advice       & The recommendations are too vague or too broad                                                       & Request the recommendations more specific  \\
Other Topic & The content of the sentence has nothing to do with the topic of medical consultation & Emphasis on medical consultation scenarios \\
Demand                 & Order or require the patient to complete actual physical actions                                     & Emphasis on medical consultation scenarios \\
\bottomrule
\end{tabular}%
}
\end{table}

\section{Results}
\subsection{Overview}
The purpose of introducing the evaluation framework is to automatically test the capabilities of doctor LLMs in diagnostic dialogues, such as gathering patient symptom information and providing effective recommendations. To more accurately assess the behavior of the LLMs in the AIE framework, we explicitly define the action space of the doctor LLMs as well as the corresponding requirements for patient simulator responses. By comparing the predefined requirements with the behaviors of LLMs during the interaction process, we can assess each interaction between the doctor LLMs and the patient simulator across multiple rounds of interaction. 

As shown in Table~\ref{tab:action and response definition}, we pre-define 10 action types of the doctor LLMs and the corresponding response requirements. 
The action space primarily contains two types of actions: inquiries and advice, corresponding respectively to the two stages of medical consultation, information gathering, and diagnosis, thus ensuring the comprehensiveness of the action space. Building on this, we further define several special action types. Initialization and conclusion types represent the start and end of the consultation dialogue, respectively. At the beginning of the consultation, patients are required to provide their chief complaints and needs to initiate a multi-round dialogue. At the end of the consultation, no response is required from the patient. Additionally, to minimize hallucinations in the model during multi-round interactions, we have defined two extra types, enabling the simulated patient to alert the physician model when hallucinations occur, thereby ensuring the smooth progression of the multi-round dialogue. Notably, the actions of inquiry and advice during the consultation process are crucial for evaluating the doctor LLMs. Consequently, we have divided these two main behaviors into three subcategories: effective, ineffective, and ambiguous. The effective category refers to inquiries and recommendations that elicit information about the patient's symptoms and examination results. The ineffective category denotes actions that fail to obtain patient information, while the ambiguous category pertains to inquiries and recommendations that are overly broad. The case study of the consultation conversation and the action categories are shown in Figure~\ref{fig: case_study}. The details of the action definition are further discussed in Section~\ref{section: state definition}. Based on the predefined action space, we can assess the performance of the patient simulator and the doctor LLMs in the multi-turn consultation conversations.

In this section, we first analyze the performance of the SAPS in multi-turn interactions to ensure its effective engagement with doctor LLMs. Subsequently, the SAPS and doctor LLMs' diagnostic dialogues are evaluated and scored using three types of methods: human evaluation and two forms of automatic evaluation, including GPT-4 evaluation~\cite{liu-etal-2023-g} and automated metric evaluation. We then analyze the correlation among the results of these three methods to validate the appropriateness of indicators. Finally, the performance of the doctor LLMs in multi-turn diagnostic dialogues is examined on the publicly available benchmarks to demonstrate the scalability of the AIE framework. This comprehensive approach underscores the efficacy of the framework in assessing doctor LLMs within the context of clinical consultations.

\subsection{Datasets}
As there is no prior work that uses AIE to assess the performance of doctor LLMs, we have developed two test sets. The first set aims to validate the effectiveness of the SAPS framework, while the second is for the AIE framework. We will outline the details of these two datasets below.
% In the study, the dataset utilized for evaluating the medical LLMs comprises two primary components. The first component consists of a collection of 50 authentic case reports and their corresponding diagnostic outcomes, which were gathered from a hospital setting. The second component involves a subset of data samples that were extracted from four publicly accessible medical multiple-choice datasets for testing purposes, including MedQA, MedMCQA, PubMedOA, and MMLU clinical topics datasets. Integrating real-world clinical data with diverse academic resources can provide a comprehensive assessment of the model's capabilities.
\subsubsection{Patient Simulator Test Sets}
%We randomly selected 50 real hospital cases as patient information. Utilizing GPT-4 to role-play both doctor and patient, we constructed 10 rounds of dialogue for each case. Depending on the dialogue context, GPT-4 can generate question-answer pairs based on predefined doctor LLM action types. Then human reviewers are hired to adjust the answer and action type of each question. This process created 4000 test questions in total, which can assess the response of patient simulators to different actions over various context lengths and effectively evaluate their interactive performance in consultation dialogues.
%We selected 50 real hospital cases randomly to represent patient information. Using GPT-4, we simulated the roles of both doctor and patient, constructing 10 rounds of dialogue for each case. Depending on the context of the dialogue, GPT-4 can generate question-answer pairs based on pre-set doctor LLM action types. Afterward, human reviewers are employed to refine the answer and action type for each question. This process yielded a total of 4000 test questions. These can be used to gauge how patient simulators respond to different actions across various context lengths and to effectively assess their interactive performance in consultation dialogues.
We randomly select 50 real hospital cases to represent patient information. We use GPT-4 to simulate the roles of both doctor and patient, creating 10 rounds of dialogue for each case. Depending on the dialogue context, GPT-4 can generate question-answer pairs based on preset doctor LLM action types. Human reviewers then refine the answer and action type for each question. This process produces 4000 test questions in total. These can evaluate how patient simulators react to different actions across various context lengths and effectively measure their interactive performance in consultation dialogues.

\subsubsection{Doctor LLMs Test Sets}
%We created two datasets to test doctor LLMs with the use of real cases and public datasets. The first dataset, called HospitalCases, consists of 50 real hospital cases not overlapping with the patient simulator test set. Each case includes patient information, examination results, and diagnoses. We used GPT-4 to generate four similar but distinct diseases as distractors, formatted as multiple-choice questions. The second dataset, MedicalExam, comprises 150 cases selected from five public datasets with lengthy patient information and diagnosis-related questions.
We establish two datasets to evaluate doctor LLMs using real cases and public datasets. The first dataset, \emph{HospitalCases}, includes 50 real hospital cases that do not overlap with the patient simulator test set. Each case comprises patient information, examination results, and diagnoses. We utilize GPT-4 to generate four similar yet distinct diseases as distractors, which are formatted as multiple-choice questions. The second dataset, \emph{MedicalExam}, encompasses 150 cases chosen from five public clinical examination datasets, including MedQA~\citep{jin2020disease}, MedMCQA~\citep{pmlr-v174-pal22a}, MMLU~\citep{hendryckstest2021}, SelfExam~\citep{10.1371/journal.pdig.0000198}, and QMAX. These cases feature extensive patient information and diagnosis-related questions.

\begin{figure*}[!th]
%是可选项 h表示的是here在这里插入，t表示的是在页面的顶部插入
\centering
\includegraphics[width=1.0\textwidth]{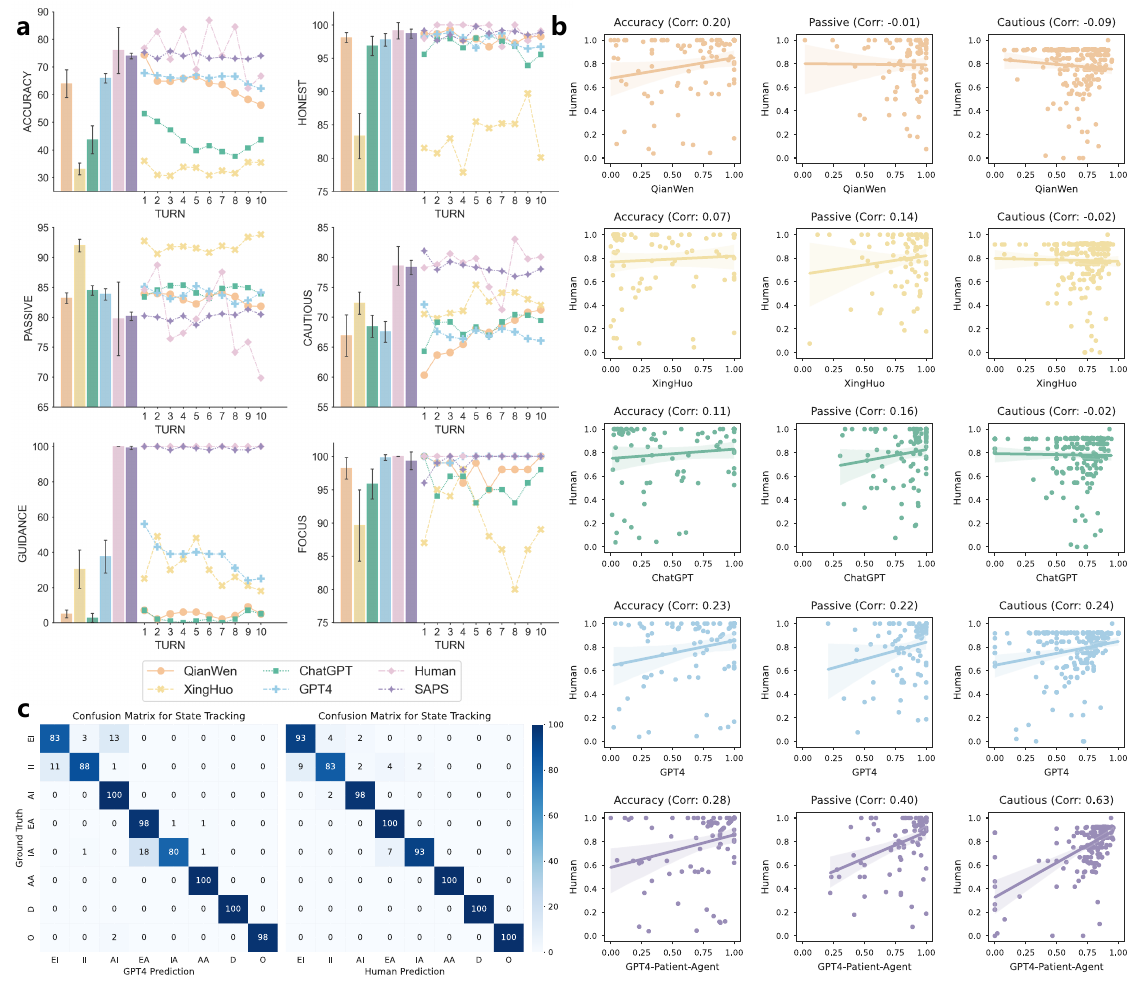}

\caption{\textbf{Results on the patient simulator test set.} We employ the six predefined patient metrics to evaluate the performance of different models and humans. \textbf{a} Change of metrics over dialogue turns. The bars and the lines in each plot describe the average scores and the relationship between the metrics and the number of dialogue turns, respectively. \textbf{b} Correlation between patient models and humans. Corr means the value of the correlation factor. \textbf{c} Confusion matrix for state tracking between patient agent and humans.} 
\label{fig: patient_result}
\end{figure*}

\subsection{Results of Patient Simulators}
In the analysis of the patient simulator's performance, we approach the evaluation from two perspectives, encompassing six dimensions in total. The first perspective focuses on how much patient information is exposed in the interaction, which is dissected into three indicators. For effective inquiries, `Accurate' measures whether the responses of the patient simulator are complete to the question of the doctor LLMs, and `Passive' assesses whether the patient simulator is over-responding. For ineffective inquiries, `Cautious' evaluates whether the patient simulator leaks the patient information without any effective question. The second perspective analyzes the behavior of the doctor LLMs. `Honest' measures whether the patient simulator denies information that does not exist in patient information, while `Guidance' and `Focus' are indicators that come into play when the doctor model's inquiries are not sufficiently professional or deviate from the consultation topic, where the patient simulator should guide or prompt the doctor LLMs. The calculations of these metrics are discussed in Section~\ref{sec: patient simulator evaluation metrics}. 

The challenge of the patient simulator test set is that the patient simulator is required to accurately identify the type of each question presented and respond according to predetermined requirements. We compare the performance of the LLM-based patient simulator with the human-acted patient, which can determine the effectiveness and accuracy of the patient simulators in consultation dialogues. As illustrated in Fig.~\ref{fig: patient_result}a, the SAPS demonstrates a performance that is closer to that of humans compared to other patient simulators. Additionally, its performance across different rounds of the dialogue is more consistent and even exhibits greater stability in scores than human participants. To analyze the performance of the patient simulators further, we conduct a comparison of the correlation between the models and humans at the instance level. As shown in Fig.~\ref{fig: patient_result}b, the SAPS demonstrates a higher correlation with human behavior in terms of the quantity of information expression. Notably, the SAPS shows improvements across all metrics when compared to the standard GPT-4 model. This suggests that the SAPS is more effective in replicating human-like interactions in the medical context. Additionally, Fig.~\ref{fig: patient_result}c presents the confusion matrix of the SAPS and human participants in the classification of the doctor LLM actions. It is found that the accuracy of the SAPS is slightly worse than that of the humans, which is still comparable and acceptable. In summary, the SAPS demonstrates superior stability and a higher correlation with human behavior than other patient simulators in the test set. The results ensure that the SAPS can reliably play the role of a patient, facilitating effective and realistic interactions with the doctor LLMs.

\begin{figure*}[th]
%是可选项 h表示的是here在这里插入，t表示的是在页面的顶部插入
\centering
\includegraphics[width=1.0\textwidth]{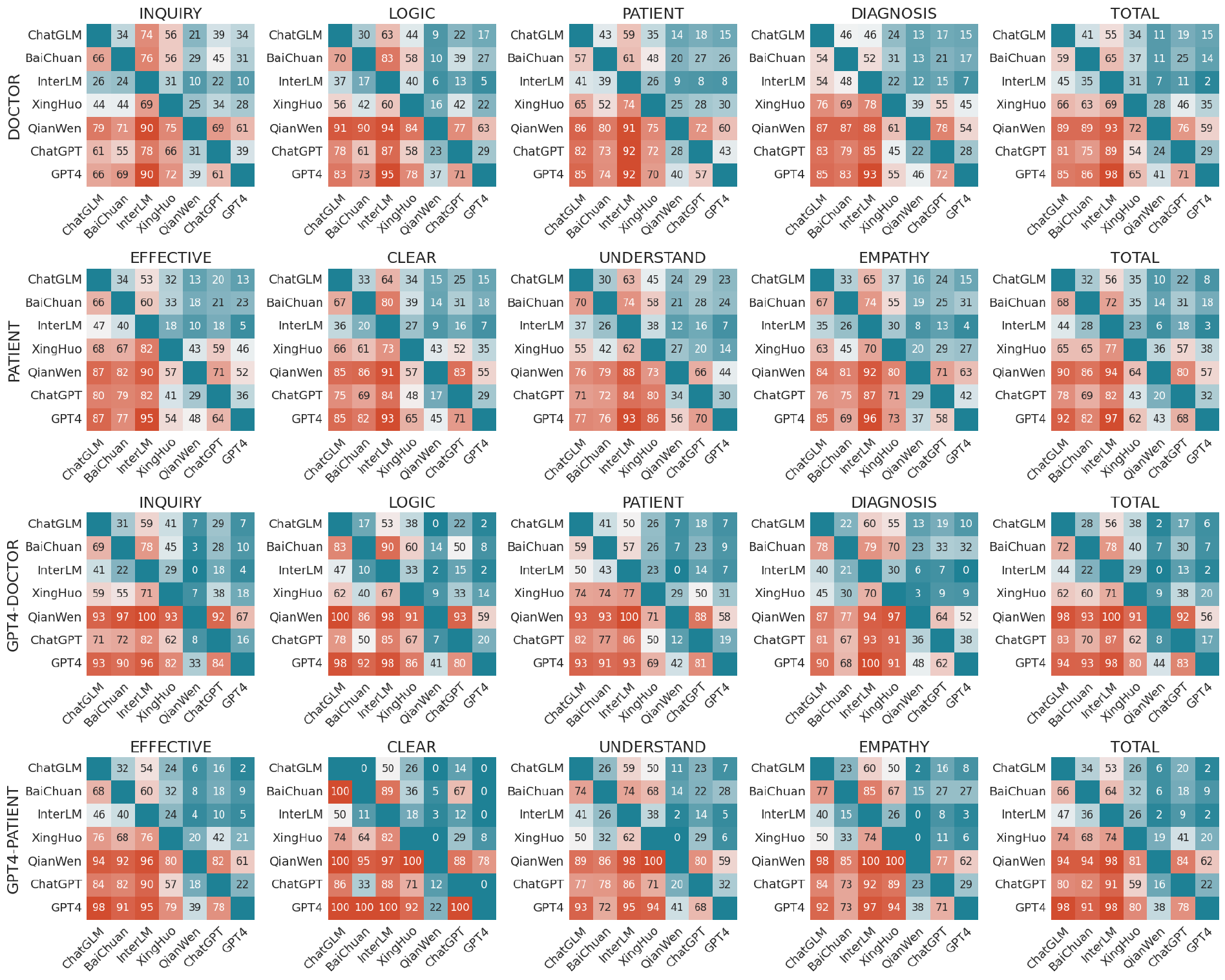}

\caption{\textbf{Success rate without tie in comparative evaluation of doctor LLMs.} The first and second rows show the results of the human evaluation from the perspective of the doctor and patient, respectively. The third and fourth rows show the results of the GPT-4 evaluation from the perspective of the doctor and patient.}
\label{fig:pair_comparison}
\end{figure*}

\begin{table}[!th]
\centering
\label{table: doctor auto}
\resizebox{\textwidth}{!}{%
\begin{tabular}{cccccccc}
\hline
\textbf{MODEL} & \textbf{ChatGLM3} & \textbf{InternLM} & \textbf{BaiChuan} & \textbf{XingHuo} & \textbf{QianWen} & \textbf{ChatGPT} & \textbf{GPT-4} \\
\hline
\textbf{DIAGNOSIS}  & 48.00$\pm$7.14 & 38.00$\pm$6.93 & 44.00$\pm$7.09 & 34.00$\pm$6.77 & 56.00$\pm$7.09 & 60.00$\pm$7.00 & 64.00$\pm$6.86 \\
\textbf{COVERAGE}    & 25.55$\pm$1.95 & 22.37$\pm$1.83 & 23.84$\pm$2.24 & 34.27$\pm$2.97 & 33.82$\pm$2.46 & 31.33$\pm$2.51 & 38.69$\pm$2.75 \\
\textbf{INQUIRY\_ACC}      & 22.75$\pm$3.11 & 15.36$\pm$3.47 & 11.64$\pm$1.92 & 31.38$\pm$3.80 & 24.80$\pm$2.94 & 27.22$\pm$3.43 & 22.52$\pm$2.54 \\
\textbf{INQUIRY\_SPECIFIC} & 72.32$\pm$3.35 & 49.79$\pm$5.63 & 62.51$\pm$2.79 & 76.58$\pm$2.22 & 94.33$\pm$1.42 & 82.37$\pm$3.31 & 90.60$\pm$2.43 \\
\textbf{INQUIRY\_LOGIC}    & 30.00$\pm$2.29 & 25.82$\pm$1.97 & 26.94$\pm$2.61 & 35.00$\pm$2.55 & 37.68$\pm$2.62 & 35.22$\pm$2.61 & 41.67$\pm$2.71 \\
\textbf{ADVICE\_ACC} & 13.00$\pm$3.91 & 13.76$\pm$4.07 & 8.33$\pm$3.57  & 20.17$\pm$5.41 & 24.80$\pm$4.71 & 15.67$\pm$3.30 & 17.83$\pm$4.37 \\
\textbf{ADVICE\_SPECIFIC}  & 27.50$\pm$5.75 & 32.83$\pm$5.62 & 18.33$\pm$4.65 & 50.83$\pm$6.95 & 55.97$\pm$5.29 & 29.67$\pm$3.64 & 48.50$\pm$5.93 \\
\textbf{DISTINCT}    & 66.66$\pm$1.75 & 62.22$\pm$3.14 & 86.02$\pm$0.47 & 65.09$\pm$2.51 & 82.64$\pm$0.55 & 74.60$\pm$1.25 & 78.19$\pm$0.48 \\
\textbf{AVG\_TURN}   & 9.82$\pm$0.08  & 9.32$\pm$0.23  & 9.06$\pm$0.20  & 8.72$\pm$0.21  & 9.28$\pm$0.14  & 9.18$\pm$0.19  & 9.50$\pm$0.13  \\
\textbf{AVG\_LEN}    & 48.26$\pm$2.98 & 46.24$\pm$2.43 & 39.44$\pm$2.40 & 87.00$\pm$6.62 & 83.88$\pm$4.66 & 55.80$\pm$2.56 & 75.62$\pm$3.43 \\
\hline
\end{tabular}%
}
\caption{Results of the automatic metrics on HospitalCases. The results are shown in the formation `mean $\pm$ standard error.'}
\end{table}

\begin{figure*}[!th]
%是可选项 h表示的是here在这里插入，t表示的是在页面的顶部插入
\centering
\includegraphics[width=1.0\textwidth]{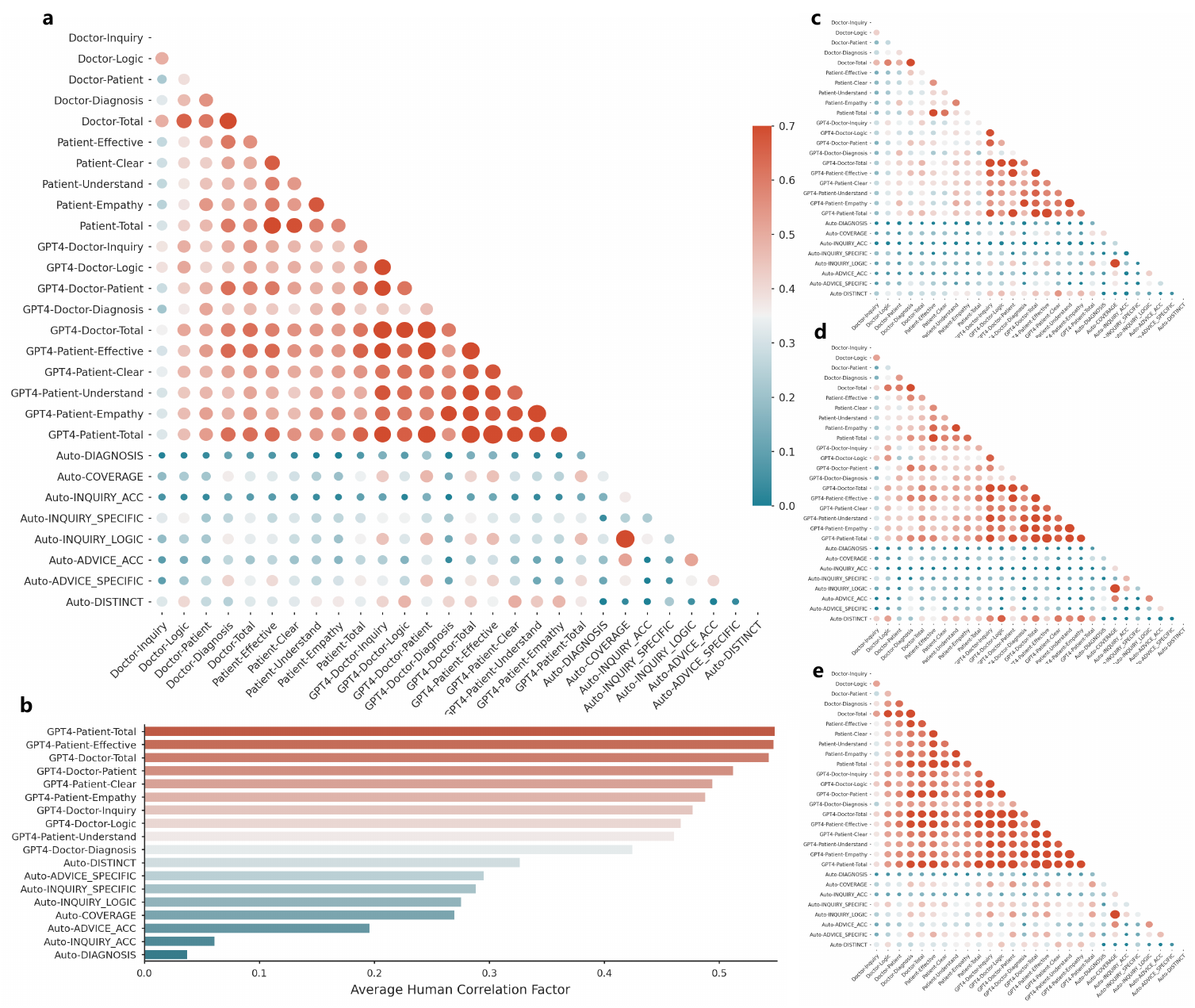}

\caption{\textbf{Analysis of the correlation between automated metrics, GPT-4, and human assessment indicators.} All indicators' correlations are tested. Considering the automated metrics are continuous, and GPT-4 and human assessments are ordinal, the Spearman correlation coefficient is used to calculate the correlation between different indicators. \textbf{a} assesses correlation across all test data. \textbf{b} the average human correlation coefficient between automated metrics and GPT-4 assessments. \textbf{c}-\textbf{e} explore correlations within specific subsets: \textbf{c} both models in the comparison are closed-source, \textbf{d} both models in the comparison are open-source, \textbf{e} one model is open-source and the other is closed-source.}
\label{fig: metric_corr}
\end{figure*}

\subsection{Comparative Evaluation on HospitalCases}
Based on the patient simulation validation, we conduct a comparative evaluation through multi-round diagnostic dialogues, which are generated by interaction between the doctor LLMs and the SAPS. In line with the criteria proposed by the UK General Medical Council Patient Questionnaire~(GMCPQ) and principles published in reviews of the consensus for best
practices for patient-centered communication (PCCBP) in medical interviews~\citep{king2013best}, we establish five metrics each from the perspectives of the doctor and patient to score the consultation dialogues. 

To avoid the instability caused by absolute scoring, we employ comparative evaluation to assess the consultation dialogues. Specifically, the dialogues of different doctor LLMs for the same case are compared pairwise to determine the more effective LLMs. We hire three medical students and normal people to play the role of doctor and patient, respectively, and choose their preference from the perspective of each metric. Additionally, we also employ various prompts to enable GPT-4 to conduct automatic evaluations from the perspectives of both the doctor and patient. 

We calculate and compare the success rates of different models in diagnosing and managing the cases accurately. The results are shown in Fig.~\ref{fig:pair_comparison}. 
The performance of QianWen stands out as the best among the models, showing the highest win rate in nearly all metrics compared to others. GPT-4 follows, exhibiting strong performance but not as dominant as QianWen. The weakest model appears to be InternLM. Overall, open-source smaller models underperform compared to closed-source larger models, aligning with findings from previous evaluations. Regarding the assessment perspectives of humans and GPT-4, GPT-4's evaluations are more polarized, indicating more extreme outcomes in terms of wins and losses within its assessment dimensions.

\subsection{Automatic Metrics Evaluation on HospitalCases}
In addition to comparative assessments, the study also involves evaluations of multi-turn diagnostic dialogues based on automated metrics. The SAPS can model the types of action, allowing for a quantitative assessment of doctor LLM behavior during diagnostic interactions. The details of the calculation of the metrics are discussed in Method. The results of the automatic evaluation are illustrated in Table~\ref{table: doctor auto}. "COVERAGE" measures the proportion of patient information recalled during the consultation, with even the strongest model, GPT-4, achieving around 38\%. `INQUIRY/ADVICE ACC' reflects the effectiveness of inquiries and advice in eliciting patient information. In contrast, "INQUIRY/ADVICE SPECIFIC" measures the specificity of inquiries and advice, distinguishing them from vague or broad questions. It's noted that the models score higher in "INQUIRY SPECIFIC" compared to "ADVICE SPECIFIC," possibly due to developers limiting the model's ability to offer direct advice for safety reasons. Additionally, there's no direct correlation between "COVERAGE," "DIAGNOSIS," and "INQUIRY/ADVICE ACC," as they measure different aspects of the diagnostic process: information recall, the effectiveness of recalled information for diagnosis, and the behavior of the model in interaction, respectively.

\subsection{Metrics Correlation Analysis}
In this section, we integrate a comprehensive set of evaluation metrics, combining human assessments from two perspectives, GPT-4 assessments from two perspectives, and various automated metrics, amounting to a total of 28 dimensions. The analysis of the correlations among these metrics is pivotal. Fig.~\ref{fig: metric_corr}a displays the interrelationships between all these metrics. From an inter-group correlation perspective, there is a significant correlation between GPT-4 annotations and human annotations. Although the corresponding metrics are not exact matches, GPT-4 shows a high degree of correlation with human judgment overall, suggesting its potential to substitute human annotations. The correlation between automated metrics and human annotations is relatively lower, with DISTINCT showing the highest overall correlation with human annotations, indicating that dialogue fluency remains a key factor in evaluating diagnostic dialogues. In terms of intra-group correlation, the metrics within the human and GPT-4 annotations show greater correlation, suggesting an overlap in these indicators. In contrast, automated metrics display less intra-group correlation, indicating their ability to independently evaluate different dimensions of the diagnostic dialogue. In addition, Fig.~\ref{fig: metric_corr}b provides a more detailed comparison of the average correlation coefficients between automated metrics, GPT-4 assessments, and human indicators. It reveals that all GPT-4 indicators have higher average human correlation coefficients than automated evaluation metrics, with the highest being scores from the perspectives of doctors and patients. Among automated metrics, DISTINCT has the highest correlation, followed by the ratio of specific inquiries and advice in the diagnostic dialogue. The lowest correlation is observed with the accuracy of diagnostic outcomes.

\begin{figure*}[!t]
%是可选项 h表示的是here在这里插入，t表示的是在页面的顶部插入
\centering
\includegraphics[width=1.0\textwidth]{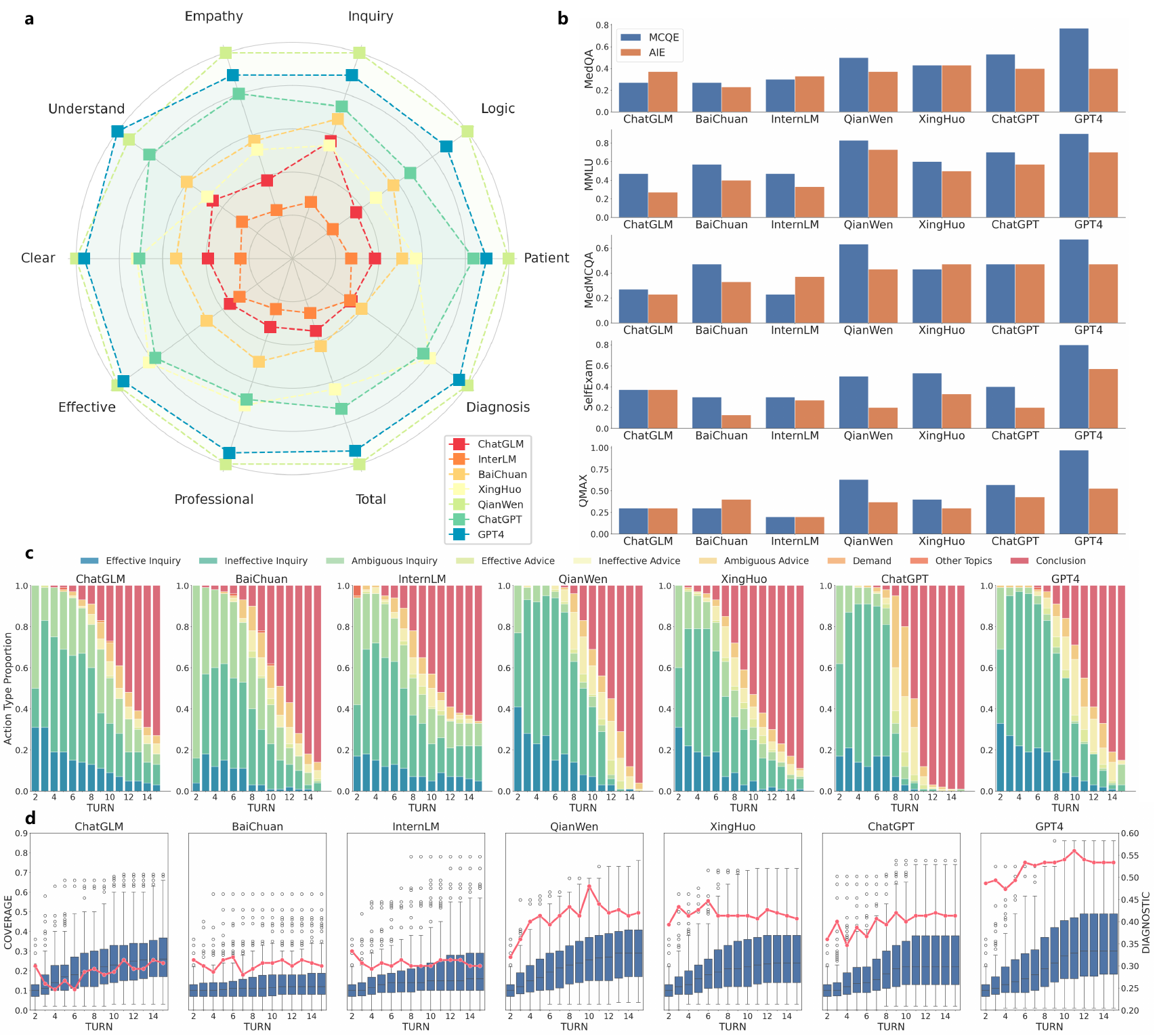}

\caption{\textbf{Analysis of the evaluated doctor LLMs on MedicalExam datasets.} \textbf{a} The average win rate of each metric within the GPT-4 evaluation. We use the best results as the maximum value for each metric in the radar chart. \textbf{b} Comparison of the diagnostic score between the Multiple-Choice Question Evaluation~(MCQE) and AIE among five subsets. \textbf{c} The proportion of the action categories within each turn. \textbf{d} The coverage and diagnostic scores depend on different consultation conversation lengths. The red lines connecting circular red dots represent diagnostic scores, while the blue box plots indicate coverage scores. } 
\label{fig: further_analysis}
\end{figure*}

\subsection{Correlation Analysis among Different Subsets}
%Considering the impact of model types on the correlation among metrics, the study categorizes the seven tested models into two groups: closed-source models, which are typically more powerful with undisclosed parameter sizes, and open-source models, which have publicly available parameters but generally smaller sizes and lower performance. Based on these distinctions, comparisons are made within three subsets: both models being closed-source, both being open-source and one of each. 
This study classifies the seven tested models into two categories based on the correlation among metrics and model types: closed-source models, which are typically more powerful but have undisclosed parameter sizes, and open-source models, which have publicly accessible parameters but are usually smaller and less efficient. Comparisons are made within three subsets: both models are closed-source, open-source, or one of each.

Figure~\ref{fig: metric_corr}c-e displays the correlations of metrics within these comparative subsets, offering insights into how model origin influences metric correlations. The analysis across different subsets of models shows varying levels of correlation among human annotations, GPT-4 evaluations, and automated metrics. 
In closed-source model comparisons (Fig.~\ref{fig: metric_corr}c), human evaluation metrics exhibit low correlation, suggesting these metrics independently assess different capabilities of the models. GPT-4 metrics show higher redundancy, indicating less distinction in evaluation aspects. In addition, the lower correlation between GPT-4 evaluations and human assessments in the subset of high-performance models suggests that GPT-4 might not yet fully replace human evaluation when comparing top-tier models. This underscores the importance of human insight in nuanced evaluation scenarios involving highly proficient models.
In Fig.~\ref{fig: metric_corr}d, which examines the second subset of comparisons (open-source models), there is a relatively higher correlation between human and GPT-4 indicators, with the automated DISTINCT metric showing strong relevance across evaluation metrics. This pattern emphasizes the importance of dialogue fluency when assessing lower-performing models, highlighting its key role in distinguishing and understanding the conversational capabilities of these models.
%In Fig.~\ref{fig: metric_corr}e, which compares a mix of open-source and closed-source models, there is a notably high correlation among automated metrics, human evaluations, and GPT-4 assessments. This increased correlation is likely due to the clear performance gap between the models within this subset, allowing for a more straightforward differentiation of model capabilities. As a result, the metrics exhibit greater alignment, reflecting a consensus on the relative strengths and weaknesses of the models compared, highlighting the distinct evaluative outcomes when contrasting models of varying origins and capabilities.
In Fig.~\ref{fig: metric_corr}e, which compares a combination of open-source and closed-source models, there is a significant correlation among automated metrics, human evaluations, and GPT-4 assessments. This heightened correlation is likely attributed to the clear performance difference between the models in this group, enabling easier differentiation of model capabilities. Consequently, the metrics show a higher alignment, indicating a consensus on the relative strengths and weaknesses of the compared models. It underscores the unique evaluation results when comparing models of different origins and capabilities.

\begin{table}[!th]
\centering
\label{table: doctor medicalexam}
\resizebox{0.95\textwidth}{!}{%
\begin{tabular}{cccccccc}
\hline
\textbf{MODEL} & \textbf{ChatGLM3} & \textbf{InternLM} & \textbf{BaiChuan} & \textbf{XingHuo} & \textbf{QianWen} & \textbf{ChatGPT} & \textbf{GPT-4} \\
\hline
\textbf{DIAGNOSIS}   & 30.67$\pm$3.78 & 30.00$\pm$3.75 & 30.00$\pm$3.75 & 15.33$\pm$2.95 & 30.67$\pm$3.78 & 41.33$\pm$4.03 & 53.33$\pm$4.09 \\
\textbf{COVERAGE}    & 27.95$\pm$1.25 & 15.36$\pm$0.96 & 21.16$\pm$1.25 & 28.08$\pm$1.40 & 31.29$\pm$1.42 & 27.09$\pm$1.44 & 33.89$\pm$1.63 \\
\textbf{INQUIRY\_ACC}      & 21.88$\pm$1.46 & 11.94$\pm$1.43 & 16.22$\pm$1.81 & 21.44$\pm$1.61 & 23.98$\pm$1.54 & 17.69$\pm$1.39 & 23.68$\pm$1.63 \\
\textbf{INQUIRY\_SPECIFIC} & 68.62$\pm$1.43 & 53.98$\pm$1.54 & 67.72$\pm$2.42 & 77.30$\pm$1.64 & 91.80$\pm$0.93 & 85.27$\pm$1.14 & 91.95$\pm$1.04 \\
\textbf{INQUIRY\_LOGIC}    & 41.27$\pm$0.82 & 28.78$\pm$0.91 & 35.61$\pm$1.03 & 40.48$\pm$1.02 & 42.59$\pm$0.90 & 40.51$\pm$1.07 & 42.74$\pm$0.99 \\
\textbf{ADVICE\_ACC} & 3.17$\pm$1.36  & 6.00$\pm$1.64  & 6.49$\pm$1.51  & 5.61$\pm$1.66  & 10.97$\pm$2.08 & 7.72$\pm$1.76  & 11.72$\pm$2.15 \\
\textbf{ADVICE\_SPECIFIC}  & 28.72$\pm$3.30 & 42.22$\pm$3.33 & 34.00$\pm$3.19 & 54.78$\pm$3.44 & 72.17$\pm$2.63 & 55.33$\pm$2.84 & 66.33$\pm$3.04 \\
\textbf{DISTINCT}    & 77.51$\pm$0.75 & 95.53$\pm$0.34 & 86.48$\pm$1.16 & 86.72$\pm$0.74 & 83.76$\pm$0.26 & 80.59$\pm$0.32 & 79.14$\pm$0.32 \\
\textbf{AVG\_TURN}   & 12.13$\pm$0.22 & 11.54$\pm$0.22 & 11.41$\pm$0.26 & 10.57$\pm$0.25 & 11.89$\pm$0.17 & 10.35$\pm$0.10 & 11.89$\pm$0.19 \\
\textbf{AVG\_LEN}    & 69.41$\pm$3.76 & 42.94$\pm$1.32 & 42.15$\pm$1.54 & 57.86$\pm$2.64 & 60.40$\pm$1.59 & 29.95$\pm$0.70 & 63.62$\pm$1.49 \\
\hline
\end{tabular}%
}
\caption{Results of the automatic metrics on MedicalExam. The results are shown in the formation `mean $\pm$ standard error.'}
\end{table}

\subsection{Automatic Evaluation on MedicalExam}
We further test doctor LLMs on the MedicalExam dataset. In Fig.~\ref{fig: further_analysis}a, the average win rate of doctor LLMs across various metrics in a GPT-4 evaluation is displayed, showing Qianwen outperforming in almost all metrics, while other models like ChatGLM and XingHuo excel in inquiry and diagnosis, respectively. Table~\ref{table: doctor medicalexam} presents results from the automated metric evaluation. It is found that score values are lower than the ones in HospitalCases, which indicates that the MedicalExam is more complex and challenging. Besides, we also found that the diagnostic result is directly contained in the patient information of HospitalCases, while MedicalExam requires further reasoning based on the collected information, which makes it harder to choose the correct diagnostic result.

\subsection{Evaluation Format Analysis}
Figure~\ref{fig: further_analysis}b shows the comparison of the diagnostic score between the Multiple-Choice Question Evaluation~(MCQE) and AIE among five subsets of MedicalExam. For the MCQE, the doctor LLMs can directly make the decision based on all the patients' information. It is observed that diagnostic scores of AIE are generally lower than those from MCQE because the diagnosis in AIE is based on information collected by the doctor model itself, which often is less comprehensive than the whole patient information. Surprisingly, in some cases, AIE diagnostic scores exceed those from MCQE, as seen with ChatGLM on MedQA~(AIE 0.37 vs. MCQE 0.27), XingHuo on MedMCQA~(AIE 0.47 vs. MCQE 0.43), and BaiChuan on QMAX~(AIE 0.40 vs. MCQE 0.30). This phenomenon is not observed with simpler datasets like MMLU or with models possessing strong reasoning abilities like GPT-4. Poor-performing models may not fully utilize complete patient information for reasoning, whereas interactive dialogue can enhance diagnostic capabilities. However, insufficient information collection during interaction directly impacts the diagnostic scores for stronger models, indicating a need for improving interaction capabilities in high-performing models.

\subsection{Turn Analysis}
Figure~\ref{fig: further_analysis}c shows the proportion of the actions. Similar to real doctors, doctor LLMs initiate consultations with simulated patients by asking questions to gather information before offering advice and diagnosis. Notably, a significant proportion of the initial inquiries are ambiguous. With feedback from the simulated patient, models with better performance quickly reduce the frequency of ambiguous inquiries. In contrast, while models with poorer performance also improve, they remain a portion of ambiguous inquiries, indicating a difference in the ability to refine questioning based on interaction quality. 

Figure~\ref{fig: further_analysis}d shows the coverage score and the diagnostic score with different lengths of the conversation. As dialogues progress, the proportion of patient information collected increases. The distribution of dots outside the box plots indicates significant variance in the difficulty of information collection across different cases. It is worth noting that diagnostic scores do not strictly correlate with the coverage score, suggesting that the helpfulness of collected information to the final diagnosis may be minimal or that the complexity of multi-round dialogues impacts model reasoning. For stronger models like GPT-4 and QianWen, there's a stronger correlation between diagnostic and coverage scores.

\section{Discussion}
Current medical benchmarks for LLMs predominantly assess medical knowledge through standardized examination questions. However, these conventional tests fail to fully evaluate LLMs in clinical application scenarios, potentially impeding the development and practical implementation of medical LLMs. Despite recent strides in medical LLM interactions, there remains a significant gap in the comprehensive evaluation of interaction rationality and definitions~\cite{johri2023testing,chen2023llm}.

In this study, we introduce the Artificial Intelligence Evaluation (AIE) framework, an innovative, cost-effective solution that leverages role-playing to simulate clinical interactions. This approach not only mimics real-life scenarios but enhances them, setting a new benchmark for testing the capabilities of LLMs. We meticulously define the action space for doctor LLMs and establish specific response requirements for our patient simulator, ensuring that the interactions are both meaningful and beneficial for patient care. Our analysis led to the identification of 10 distinct action-response types, optimizing the interaction process. A key innovation is the introduction of an 'ambiguous' category to address the frequent issue of LLMs producing verbose yet unhelpful responses. This strategic filter significantly improves communication efficiency and encourages deeper, more meaningful engagement between doctor LLMs and the patient simulator.

A major breakthrough in our research is the development of SAPS, a sophisticated patient simulator that sets new standards for realism and performance in simulating patient responses. SAPS's ability to mimic human-like responses with remarkable accuracy suggests its potential to replace human participants in specific training scenarios, thereby revolutionizing medical training by increasing scalability and reducing human resource dependency. The exceptional performance of SAPS across various metrics, notably its stability in extended multi-round dialogues, underscores its capability to effectively manage the dynamic and complex nature of real-life clinical interactions.

Our findings highlight that SAPS's responses bear a closer resemblance to those of human patients than previous models, enhancing the realism of the simulations and suggesting its potential to streamline the training process, making it more efficient and scalable. Furthermore, SAPS's sophisticated ability to interpret the actions of doctor LLMs during consultations provides invaluable insights into refining the decision-making and communicative strategies of the simulated doctors, which are critical for improving medical training programs.

Moreover, the evaluation of the AIE framework through automated methods reveals a significant alignment with human judgments, confirming the robustness of using GPT-4 for assessing the interaction capabilities of doctor LLMs~\cite{DBLP:journals/corr/abs-2303-08774}. Correlation coefficients between GPT-4's evaluation metrics and human assessment indicators are impressively consistent, indicating that automated assessments within the AIE framework can effectively stand in for human evaluations. This alignment highlights the effectiveness of our specifically designed metrics in capturing essential aspects of interaction quality that resonate with human evaluators.

The experimental results from the AIE framework underscore a significant disparity between weak and strong models in multi-turn medical consultation dialogues. Strong models demonstrate a definitive advantage in both human evaluations and automated assessments. Additionally, the interaction process within the AIE framework offers the potential for enhancing the interactive capabilities of these models, as evidenced by the specificity and accuracy metrics in the correlation analysis. These metrics are crucial for refining diagnostic processes during consultations, thus enhancing the overall efficacy of the diagnostic task.

While this study significantly advances the validation of the AIE framework from multiple perspectives, it is important to acknowledge some inherent limitations of our research methodology. These limitations primarily stem from the observational approach used to define doctor actions and patient response criteria in our simulations, which might not capture the full scope and complexity of real-world medical interactions. As evidenced by the notable alignment of the SAPS model's responses with human behaviors within a specifically defined action space, the ongoing need to broaden and refine this space is clear as medical modeling technology evolves. Additionally, the reliability of the metrics used in our study varies; they are particularly effective in differentiating between models with significant performance differences but prove less capable when assessing models with similar performance levels. This variability in metric reliability also poses challenges in human evaluations, especially when attempting to discern fine distinctions among top-performing models.

Despite these challenges, the interactive evaluation methodology we introduce offers substantial potential for advancing research. This approach not only presents a more dynamic and realistic evaluation environment compared to traditional methods but also supports the development and application of large language models across various domains beyond healthcare. The methodology is especially beneficial in fields that require authentic scenario testing, promising extensive applicability across diverse tasks that demand high fidelity in real-world interactions. This makes it an invaluable asset for the research community, poised to enhance the utility and effectiveness of language models in numerous applications.~\cite{mehandru2024evaluating}

\section{Methods}

% \subsection{Problem Formulation}

\subsection{Evaluated Large Langauge Models}
In general, most existing LLMs can be divided into the general LLMs and the domain LLMs, which specifically refer to the medical LLMs in this paper. Preliminary experiments have shown that the instruction-following ability of many medical LLMs is compromised, causing them to fail multi-turn consultation tasks requiring complex instructions. As such, we are selecting general LLMs with strong instruction-following capabilities and bilingualism to maintain fairness in experimental comparison. For the closed-source models, ChatGPT~\citep{elmohamed}, GPT-4~\citep{DBLP:journals/corr/abs-2303-08774}, Xinghuo, and Qianwen~\citep{qwen} are selected. For the open-source models, Internlm-20b~\citep{2023internlm}, Baichuan-13b~\citep{yang2023baichuan}, and ChatGLM3-6b~\citep{du2022glm} are selected. Specifically, we use the version \texttt{gpt-3.5-turbo-1106} for ChatGPT and \texttt{gpt-4-1106-preview} for GPT-4. For XingHuo, we use the version of \texttt{generalv3}. For QianWen, we use the version of \texttt{qwen\_max}. For the closed-source models, we use the \texttt{ChatGLM3-6B} available at {\url{https://github.com/THUDM/ChatGLM3}} for ChatGLM3-6b, \texttt{Baichuan-13B-Chat} available at {\url{https://github.com/baichuan-inc/Baichuan-13B}} for BaiChuan-13b, and \texttt{InternLM2-Chat-20B} available at {\url{https://github.com/InternLM/InternLM}} for InternLM-20b.

\subsection{State Aware Patient Simulator}
During the consultation conversation, the patient agent is required to provide tailored responses according to different types of questions. However, directly prompting LLMs makes it difficult to control the behavior of the patient simulator precisely. To enhance the precision of patient simulators in consultation dialogues, we introduce a State Aware Patient Simulator (SAPS). SAPS consists of three integral components: a state tracker that identifies and classifies the current action state of the doctor LLMs, a memory bank that stores various information and selects the appropriate memory according to the state, and a response generator that produces contextually relevant replies. This architecture allows SAPS to respond to a wide range of inquiries adaptively by understanding the consultation's dynamic context, thereby generating responses that are both accurate and contextually appropriate for the evolving dialogue.

\subsubsection{States definition and response requirements}
\label{section: state definition}
%In our approach, we distinguish the types of actions performed by doctors during medical consultations based on two key perspectives: the nature of the doctor's actions and their effectiveness. The first perspective categorizes the actions into distinct types that encompass the range of interactions a doctor might engage in during a patient consultation. The second perspective assesses the effectiveness of these actions, which focuses on whether the doctor's actions successfully elicit relevant information and provide applicable advice. Therefore, we predefined 10 types of state and the corresponding requirements in the consultation conversation as follows: 
Our approach categorizes the types of actions performed by doctors during medical consultations from two key perspectives: the nature of the actions and their effectiveness. The first perspective classifies the actions into distinct categories that cover the array of interactions a doctor might have during a patient consultation. The second perspective evaluates the effectiveness of these actions, concentrating on whether they successfully gather pertinent information and offer suitable advice. Consequently, we predefine 10 categories of states and their corresponding requirements in the consultation conversation as follows:

\textbf{Initialization:} Actions initiating the consultation between the doctor LLM and the patient simulator. In this phase, the doctor LLMs typically begin the consultation with a consistent approach, such as `Hello, I'm your doctor. How can I help you today.' The patient simulator should briefly describe their main symptoms and primary concerns, offering a clear but brief overview of the health issues without delving into exhaustive detail. 

\textbf{Inquiry:} This category involves the doctor LLMs asking the patient for information, which can be divided into three subclasses based on the effectiveness of the question. The first is effective inquiries, including questions that are specific and correctly guide the patient to disclose their physical condition or relevant information. In this case, SAPS is designed to respond with the corresponding part of the patient information relevant to the question. This requirement maintains the accuracy of the patient's information without altering its original meaning through rephrasing. The second is ineffective inquiries. This type refers to questions that are specific but fail to gather the patient information, such as asking for information that is not recorded in the patient information. In this case, the SAPS should indicate a denial or express uncertainty regarding the information requested to avoid providing misleading or fabricated information. The final is ambiguous inquiries, which represent questions without a specific focus and are overly broad. For this type of question, the patient should require the doctor to ask more specifically instead of directly providing a comprehensive response that covers a wide range of information. In our experiments, we observe that the doctor model could elicit more information from the patient simulator by repeatedly asking vague questions. However, such behavior does not align with the original intent of our testing framework, which aims to encourage the model to ask more specific and professional questions. This requirement is designed to prevent models from taking shortcuts within the framework.

\textbf{Advice:} %This type encompasses recommendations given by the doctor model. Considering that the simulated consultation interaction cannot involve actual medical examinations and obtain the results in real-life consultations, we treat the advice action of the doctor LLMs as inquiries actions, which enable a comprehensive evaluation of the doctor model's ability to provide actionable advice within the constraints of the simulation. Therefore, it is assumed that the SAPS can reply with advice (like some kind of medication treatment or medical examination) if its results are contained in patient information. Similar to inquiries, the advice category can also be divided into three subclasses, including effective advice, ineffective advice, and ambiguous advice. Effective advice refers to suggestions that the results are contained in patient information and SAPS needs to respond with the corresponding part of patient information. The second is ineffective advice, where the patient's information does not contain the results, and the patient should express negation. The final one is ambiguous advice that lacks specific practices. Similar to the requirement of the ambiguous inquiry, SAPS also needs to require doctor LLMs to ask for more specifics.
This category includes recommendations made by the doctor model. Given that simulated consultations can't involve actual medical examinations or obtain real-life results, we treat the doctor model's advice as inquiry actions. This allows for a thorough evaluation of the model's ability to provide actionable advice within the simulation's constraints.
We assume the SAPS can respond with advice if the results are included in the patient's information, such as medication treatment or medical examination. Similar to inquiries, advice can also be divided into three subclasses: effective advice, ineffective advice, and ambiguous advice.
Effective advice refers to suggestions where the patient's information contains the results, and SAPS needs to respond with the corresponding information. Ineffective advice is when the patient's information does not contain the results, and the patient should express negation. Ambiguous advice lacks specific practices, and like an ambiguous inquiry, SAPS needs to ask the doctor model for more specifics.

\textbf{Demand:} %This category involves actions of doctor LLMs asking the patient to perform certain physical actions that are not feasible during an online consultation. SAPS needs to refuse the request and remind the doctor that the medical consultation is online and is unable to complete physical actions.
This category includes instances where doctors with LLMs ask patients to perform physical actions that are impossible during an online consultation. SAPS should deny these requests and remind the doctor that the consultation is online, making physical actions unfeasible.

\textbf{Other Topics:} %This category includes all actions by doctor LLMs that are unrelated to the diagnostic dialogue. It encompasses inquiries and conversations that do not pertain to the patient's symptoms or condition, deviating from the main topic of the consultation. In such scenarios, the SAPS is designed to guide the doctor back to the context of the consultation, re-emphasizing its concerns and symptoms. This ensures the dialogue remains focused on the patient's health issues, maintaining the relevance and efficiency of the consultation process.
This category includes all actions by Doctor LLMs that are not related to the diagnostic dialogue. These are inquiries and conversations that deviate from the main topic of the consultation and do not concern the patient's symptoms or condition. In these situations, the SAPS is designed to redirect the doctor back to the consultation context, re-emphasizing the patient's symptoms and concerns. This keeps the dialogue focused on the patient's health issues, ensuring the consultation remains relevant and efficient.

\textbf{Conclusion Category:} %Actions signaling the end of the consultation, including summarizing the visit, arranging a follow-up, or bidding farewell. In this case,  the patient model is not required to give a response.
Actions that signal the end of the consultation may include summarizing the visit, arranging a follow-up, or saying goodbye. In these instances, the patient model is not required to respond.

\subsubsection{State tracker}
\label{section: state tracker}
The state tracker plays a role in categorizing the actions of doctor LLMs into a predefined type. The prediction process can be divided into three steps. The first step is to classify the current action into five main categories, excluding the initial category. This is because the initial category is fixed as the initial round of the consultation dialogues. SAPS classifies the action with the use of prompt as follows:
\begin{verbatim}
In the process of medical consultation, a doctor's questions can 
be classified into five types:
    (A) Inquiry: The doctor asks the patient for medical and 
        disease-related symptom information. Generally, questions 
        with a '?' that do not belong to categories (C) or (D) are 
        included in this category.
    (B) Advice: The doctor suggests that the patient visit a 
        hospital for consultation, undergo examinations, or 
        provide certain treatment plans. Questions containing 
        the keyword 'suggestion' belong to this category.
    (C) Demand: The doctor asks the patient to perform certain 
        actions for observation, cooperation, or sensation. 
        Actions include but are not limited to opening the mouth, 
        lying on the side, standing, pressing, etc.
    (D) Other Topics: The doctor's questions do not pertain to 
        the medical consultation context and are unrelated to 
        medical diseases. This includes, but is not limited to, 
        hobbies, movies, food, etc.
    (E) Conclusion: The doctor has completed the consultation 
        and does not require a response from the patient.
        
Based on the descriptions of the above question types, please 
choose the most appropriate category for the following <Doctor 
Question>:

<Doctor Question>: {{{question}}}
Question Type: (    
\end{verbatim}

To induce the LLMs to generate valid responses, we take advantage of the \textbf{logit\_bias} technique~\citep{DBLP:journals/corr/abs-2303-13375} to enforce the model only generate the category index. If the current action falls into the categories of inquiry or advice, the second step of classification is to determine whether the action is specific or ambiguous. For inquiries and advice, different prompts are used to facilitate this distinction. For the inquiries category, the prompt is shown below:
\begin{verbatim}
<Definition>:
[Specific]: <Question> has a certain specific direction. When 
asking about symptoms, it should at least inquire about specific 
body parts, symptoms, sensations, or situations. When asking 
about examination results, it should mention specific body parts, 
specific examination items, or abnormal situations. Note that if 
it's about specific medical conditions, like medical history, 
family history, chronic illnesses, surgical history, etc., they 
are always considered [Specific]. Specifically, if the <Question> 
contains demonstrative like "these" or "this", then it is related 
to the above and should belong to the [Specific].
[Ambiguous]: <Question> such as "Where do you feel uncomfortable?" 
or "Where does it feel strange?" without any specific information
direction are considered [Ambiguous].

<Question>: {question}

Based on the <Definition>, determine whether the doctor's <Question> 
asks for [Specific] medical information from the patient or gives 
[Specific] advice. If so, directly output [Specific]. If not, output
[Ambiguous].
\end{verbatim}

For the advice category, the prompt is shown as below:
\begin{verbatim}
<Definition>:
[Specific]: <Advice> contains specific types of examinations or test
(including but not limited to X-rays, MRI, biopsy, etc.), specific 
treatment plans (including but not limited to specific surgical 
treatments, exercises, diets, etc.), specific types of medication, 
etc.

[Ambiguous]: <Advice> broadly given without any specific examination/
test, treatment plans, doctor's orders, exercises, diets, and 
medication types are considered [Ambiguous]. As long as any of the 
above information appears, <Advice> does not fall into this category.

<Advice>: {question}

Based on the <Definition>, determine whether the doctor's <Advice> 
asks for [Specific] medical information from the patient or gives 
[Specific] advice. If so, directly output [Specific]. If not, output 
[Broad].
\end{verbatim}

If the action is determined to be specific, the next step is to assess whether there is a corresponding answer within the patient information. Similarly, we also prepare two types of prompts for the inquiry and advice:
\begin{verbatim}
<Definition>:
[Relevant Information]: <Patient Information> contains information 
asked in <Question>, including descriptions of having or not having 
the symptom, as long as there's relevant content.
[No Relevant Information]: <Patient Information> does not contain 
the information asked in <Question>, and there's no relevant content 
in the information.

<Patient Information>: {patient_info}

<Question>: {question}

Based on the <Definition>, determine whether <Patient Information> 
contains relevant information asked in <Question>. If [Relevant 
Information] is present, directly output the relevant text stat-
ement, ensuring not to include irrelevant content. If [No Relevant 
Information], then directly output [No Relevant Information].
\end{verbatim}

\begin{verbatim}
<Definition>:
[Relevant Information]: <Patient Information> contains results of 
the examinations or treatment plans suggested in <Advice>, including 
any results related to the suggested examination items and treatment 
plans.
[No Relevant Information]: <Patient Information> does not contain 
results of the examinations or treatment plans suggested in <Advice>,
including no mention of relevant examination items and treatment 
plans or no corresponding results.

<Patient Information>: {patient_info}

<Advice>: {question}

Based on the <Definition>, determine whether <Patient Information> 
contains relevant information about the measures suggested in <Advice>. 
If [Relevant Information] is present, directly output the relevant 
text statement, ensuring not to include irrelevant content. If 
[No Relevant Information], then directly output 
[No Relevant Information].
\end{verbatim}

In conclusion, the predicted categories at each turn can be identified as:
 \begin{equation}
    \hat{s}_i = \Phi_{\rm STraker}(D_i, \textbf{M}_{\rm long})
\end{equation}
where $D_i$ indicates the doctor's output of the $i$ turn and $\Phi_{\rm STraker}$ is also the LLMs. 

\subsubsection{Memory bank}
Upon identifying the current action type of the doctor LLM through the state tracker, SAPS can selectively access different parts of the memory bank. This process enables the generation of SAPS responses, ensuring they are driven by and aligned with the categorized behavior type. The memory bank comprises three parts, each serving a specific function in facilitating realistic and adaptive interactions with the doctor.

Long-term Memory: This component stores the patient information $\textbf{M}_{\rm long}$ and remains consistent during the interaction. The details contained in long-term memory, such as medical history, personal health data, and possibly previous experiences with healthcare, are crucial to play the patient's role, which enables the SAPS to provide consistent and accurate information during the consultation process. In each turn, the long-term memory will extract the corresponding information relevant to the doctor's actions only when their types belong to the effective inquiry and effective advice categories. The extracted process of the long-term memory in $i$-th turn can be denoted as:
\begin{equation}
    \textbf{m}_{{\rm long}, i} = \Phi_{\rm LTM}(D_i, \textbf{M}_{\rm long}, \hat{s}_i)
\end{equation}

Working Memory: The working memory $\textbf{M}_{\rm work}$ is pivotal for the adaptive responses of the SAPS to different actions of doctor LLMs. It stores the requirements for responding to each type of action defined in Section~\ref{section: state definition}, enabling the SAPS to react appropriately to the evolving context of the consultation. Therefore, the working memory of the $i$-th turn can be denoted as:
\begin{equation}
    \textbf{m}_{{\rm work}, i} = \textbf{M}_{\rm work}[\hat{s}_i]
\end{equation}
% 大概描述一下working memory的回答方式

Short-term Memory: This part retains the history of the consultation dialogue between the doctor LLM and the SAPS. It will continually with the flow of the conversation. This enables the SAPS to reference recent interactions, ensuring continuity and relevance in the dialogue, which is essential for a coherent and meaningful conversation in a real-time consultation setting. The short-term memory in the $i$-th can be expressed as:
\begin{equation}
    \textbf{m}_{{\rm short}, i} = 
        \bigcup_{j=1}^{i-1} \{(D_j, P_j)\}
\end{equation}
where $P_j$ is the patient response in $j$-th turn of the conversations and $\textbf{m}_{{\rm short}, 0}=\emptyset$. 

\subsubsection{Response generator}
The memory information extracted from the memory bank in $i$-t turn can be noted as $\textbf{m}_i=\{\textbf{m}_{{\rm long},i}, \textbf{m}_{{\rm work},i},\textbf{m}_{{\rm short},i}\}$ and the prompt for the response generator will be organized as the context in the predefined format as below:
\begin{verbatim}
<Patient Information>: {long_term_memory}

<Requirements>: {working_memory}

The following is a conversation between a doctor and a patient. 
The patient will respond to the latest round of the doctor's 
question in the first person according to the <Requirements>. 
Note that do not output any text content in <Requirements>!

<Dialong>: 
{short_term_memory}
[Doctor]: {doctor_question}
\end{verbatim}
Then the response generator will generate the response in an autoregressive manner, where each output token is modeled depending on the previous tokens, the memory information, and the current action of the doctor LLM.
\begin{equation}
    \Phi_{\rm gen}(P_i|D_i, \textbf{m}_i) = \prod^{|P_i|}_{j}\Phi_{\rm gen}(p_{i,j}|p_{i,\textless j}, D_i, \textbf{m}_i)
\end{equation}
where $|P_i|$ and $p_{i,j}$ represents the number of tokens in the $P_i$ and the $j$-th token in the $P_i$, respectively.

\subsection{Medical LLMs Evaluation Pipeline}
In the proposed approach, the whole evaluation pipeline can be divided into two parts: multi-turn medical consultation and diagnosis. In this first part, the doctor LLMs engage in a conversational interaction to gather as much information as possible about the patient's symptoms within a predefined length of dialogue rounds. The model must efficiently and effectively probe for details, asking relevant questions and interpreting the responses to build a comprehensive understanding of the patient's condition. Here the doctor LLMs  output the sequence also in an autoregressive manner:
\begin{equation}
    \Phi_{\rm doctor}(D_i|\textbf{m}_i) = \prod^{|D_i|}_{j}\Phi_{\rm doctor}(d_{i,j}|d_{i,\textless j}, \textbf{m}_{{\rm short},i})
\end{equation}
where $|D_i|$ and $d_{i,j}$ represents the number of tokens in the $D_i$ and the $j$-th token in the $D_i$, respectively.  

The second part involves the doctor LLMs making a diagnosis based on the information collected during the multi-turn dialogue. Here, the doctor LLMs are tasked with analyzing and synthesizing the patient's responses to determine the most likely medical condition from a set of potential diagnoses. 

\subsection{Evaluation Metrics}
To accurately measure and assess the performance of the LLMs within the proposed framework, we need to redefine and set specific criteria. As such, we have proposed metrics for evaluating both the patient agent and medical LLMs. Given the patient simulation test set of size $N$, for each doctor action $D_i$, $i\in{1,2,...,N}$, its type is denoted as $s_i$, with a dialogue context $\textbf{m}_{{\rm short}, i}$, and the response of the patient simulator is denoted as $P_i$.
\subsubsection{Patient Simulator Evaluation Metrics}
\label{sec: patient simulator evaluation metrics}
The role of patient agents in medical consultations is critically influential in ensuring the fairness of evaluations for doctor models. The accuracy and relevance of the information provided by these patient agents directly impact the ability of doctor models to make appropriate diagnoses and recommendations. Based on the state definition and response requirements discussed in Section~\ref{section: state definition}, we evaluate the performance of the patient agent from 6 perspectives.

\textbf{ACCURACY}. This dimension assesses the model's ability to provide correct and relevant information in response to the patient's condition. For the question belonging to the effective inquiry and effective advice, it is expected that the reply of the patient agent can cover the corresponding part of the patient information. Therefore, we validate the accuracy with the recall of the Rouge score~\cite{lin2004rouge} between the reply and the ground truth answer. The metric score can be calculated as:
\begin{equation}
    {\rm ACCURACY} = \frac{\sum_{i=1}^{N}\mathbb{I}(s_i\in\{{\rm EI,EA}\}) \cdot \text{Recall}(P_i, \textbf{m}_{{\rm gt, i}})}{\sum_{i=1}^{N} \mathbb{I}(s_i\in\{{\rm EI,EA}\})}
\end{equation}
where $\textbf{m}_{{\rm gt, i}}$ is the ground truth extracted from the patient information corresponding to the doctor's action only when the question belongs to the EI, EA categories, where EI and EA are abbreviations for effective inquiry/advice categories, respectively. $\text{Recall}(\cdot, \cdot)$ using the implement of the recall score of the Rouge-1~\cite{lin2004rouge}.  

\textbf{HONEST}. This dimension is used to judge whether the patient agent has fabricated facts that do not exist in the patient's information. For the question belonging to the ineffective inquiry and ineffective advice, it is expected that the patient agent will express negation to avoid providing misleading or fabricated information. Only when the answer of the patient agent contains the word in the predefined negative word set is the patient honest. HONESTY measures whether the patient simulator denies information that does not exist in patient information when the actions of the doctor LLMs belong to ineffective inquiry/advice categories~(II/IA):
\begin{equation}
    {\rm HONESTY} = \frac{\sum_{i=1}^{N} \mathbb{I}(s_i\in\{{\rm II,IA}\}) \cdot \mathbb{I}(\bigvee_{h \in H} h \in P_i)}{\sum_{i=1}^{N} \mathbb{I}(s_i\in\{{\rm II,IA}\})}
\end{equation}
where $H$ represents a set of honest keywords. If any words $h\in H$ appear in the simulated patient's response, we consider the simulated patient to be honest in that answer. The symbol $\bigvee$ represents the logical OR operator, indicating that the expression is true if at least one of the conditions is satisfied. 

\textbf{FOCUS}. FOCUS measures whether the patient simulator prompts the doctor LLMs to focus on the medical consultation topic when the doctor LLMS' actions belong to the other topic categories~(OT) and whether they prompt the doctor LLMs about the online consultation scenarios when the doctor LLMS' actions belong to the demand categories~(DE):
\begin{equation}
    {\rm FOCUS} = \frac{\sum_{i=1}^{N} \mathbb{I}(s_i\in\{{\rm OT,DE}\}) \cdot \mathbb{I}(\bigvee_{f \in F} f \in P_i)}{\sum_{i=1}^{N} \mathbb{I}(s_i\in\{{\rm OT,DE}\})}
\end{equation}
where $F$ represents a set of focus keywords. If any words $f\in F$ appear in the simulated patient's response, we consider the simulated patient to be focus in that answer.

\textbf{PASSIVE}. This dimension is used to validate how much extra information is exposed when the patient agent replies to effective inquiries and advice. Lower scores indicate that the patient agent tends to answer more information compared to the corresponding answer, which makes the overall task easier. We measure the amount of additional patient information in a patient's reply by calculating the overlap between the patient's reply and the entire patient information, then subtracting the overlap between the reply and the corresponding patient information. To ensure consistency in the normalization standards for comparing two sets of information, we use precision to calculate the degree of overlap.
\begin{equation}
\begin{aligned}
    {\rm PASSIVE}  &= \frac{\sum_{i=1}^{N} \mathbb{I}(s_i\in\{{\rm EI,EA}\}) \cdot \text{Precision}(P_i, \textbf{m}_{{\rm long, i}}))}{\sum_{i=1}^{N} \mathbb{I}(s_i\in\{{\rm EI,EA}\})} \\ & - \frac{\sum_{i=1}^{N} \mathbb{I}(s_i\in\{{\rm EI,EA}\}) \cdot \text{Precision}(P_i, \textbf{m}_{{\rm gt, i}})}{\sum_{i=1}^{N} \mathbb{I}(s_i\in\{{\rm EI,EA}\})}
\end{aligned}
\end{equation}
where $\textbf{m}_{{\rm long, i}}$ is the whole patient information of the case.

\textbf{CAUTIOUS}. Similar to the Passive score, CAUTIOUS evaluates whether the patient simulator leaks the patient information when the doctor's action belongs to the II/AI. In this case, no ground truth information corresponds to the ineffective action. Therefore, any patient information included in a response is considered an act of information leakage:
\begin{equation}
    {\rm CAUTIOUS} = \frac{\sum_{i=1}^{N}\mathbb{I}(s_i\in\{{\rm EI,EA}\}) \cdot \text{Precision}(P_i, \textbf{m}_{{\rm long, i}})}{\sum_{i=1}^{N} \mathbb{I}(s_i\in\{{\rm EI,EA}\})}
\end{equation}

\textbf{GUIDANCE}. GUIDANCE measures whether the patient simulator guides the doctor LLMs to act more specifically when the doctor LLMS' actions belong to the ambiguous inquiry/advice categories~(AI,AA):
\begin{equation}
    {\rm GUIDANCE} = \frac{\sum_{i=1}^{N} \mathbb{I}(s_i\in\{{\rm AI,AA}\}) \cdot \mathbb{I}(\bigvee_{g \in G} g \in P_i)}{\sum_{i=1}^{N} \mathbb{I}(s_i\in\{{\rm AI,AA}\})}
\end{equation}
where $G$ represents a set of guidance keywords. If any words $g\in G$ appear in the simulated patient's response, we consider the simulated patient to be guidance in that answer. 

\subsubsection{Doctor LLMs Automatics Evaluation Metrics}
For the test to the doctor LLMs, assuming the size of the test set is $N$, for the $i$-th case, the patient information is $\textbf{m}_{{\rm long}, i}$, the length of the dialogue generated by the AIE is $L_i$, the ground truth diagnosis is $r_i$.

\textbf{DIAGNOSIS} DIAGNOSIS measures the accuracy of the final diagnosis task depending on the collected patient information during the consultation process:
\begin{equation}
    {\rm DIAGNOSIS} = \frac{1}{N}\sum_{i=1}^{N}\mathbb{I}(c_i==r_i)
\end{equation}
where $c_i$ is the prediction of the doctor LLMs.

\textbf{COVERAGE} COVERAGE measures how much patient information is recalled during the consultation conversation process. We first merge the information collected in the dialogue:
\begin{equation}
    \textbf{m}_{{\rm col},i} = \bigcup_{s^{i}_j \in \{{\rm EI,EA}\}}\{P_j\}
\end{equation}
where the $s^{i}_j$ indicates the categories of the doctor action in $i$-th case $j$-th turn. Note that we only consider the information collected in the action belonging to EI/EA. Then we can calculate the COVERAGE score as below:
\begin{equation}
    {\rm COVERAGE} = \frac{1}{N}\sum_{i=1}^{N}{\rm Recall}(\textbf{m}_{{\rm col},i}, \textbf{m}_{{\rm long}, i})
\end{equation}

\textbf{INQUIRY ACC} This metric measures the proportion of the EI in the inquiry actions:
\begin{equation}
    \text{INQUIRY ACC} = \frac{\sum_{i=1}^{N}\sum_{j=1}^{L_i}\mathbb{I}(s^i_j == {\rm EI})}{\sum_{i=1}^{N}\sum_{j=1}^{L_i}\mathbb{I}(s^i_j \in \{{\rm EI,II,AI}\})}
\end{equation}

\textbf{INQUIRY SPECIFIC} This metric measures the proportion of the EI/II in the inquiry actions:
\begin{equation}
    \text{INQUIRY SPECIFIC} = \frac{\sum_{i=1}^{N}\sum_{j=1}^{L_i}\mathbb{I}(s^i_j\in \{{\rm EI,II}\})}{\sum_{i=1}^{N}\sum_{j=1}^{L_i}\mathbb{I}(s^i_j \in \{{\rm EI,II,AI}\})}
\end{equation}

\textbf{INQUIRY LOGIC} This metric measures the gathering information logic of the doctor LLMs. We discovered that hospital records and examination questions follow the consultation sequence. Typically, they start with basic personal information, including gender and age, followed by chief complaints, then move on to medical and travel history, and conclude with various examination indicators and results. Therefore, a straightforward method to evaluate a model's logic in collecting patient information is by calculating the edit distance between the collected patient information and the standard patient information. In our experiment, we use Levenshtein Distance~(LD)~\cite{levenshtein1966binary} to calculate the edit distance:
\begin{equation}
    \text{INQUIRY SPECIFIC} = \frac{1}{N}\sum_{i=1}^{N}{\rm LD}(\textbf{m}_{{\rm col},i}, \textbf{m}_{{\rm long}, i})
\end{equation}

\textbf{ADVICE ACC} This metric measures the proportion of the EA in the advice actions:
\begin{equation}
    \text{ADVICE ACC} = \frac{\sum_{i=1}^{N}\sum_{j=1}^{L_i}\mathbb{I}(s^i_j == {\rm EA})}{\sum_{i=1}^{N}\sum_{j=1}^{L_i}\mathbb{I}(s^i_j \in \{{\rm EA,IA,AA}\})}
\end{equation}

\textbf{ADVICE SPECIFIC} This metric measures the proportion of the EA/IA in the inquiry actions:
\begin{equation}
    \text{INQUIRY SPECIFIC} = \frac{\sum_{i=1}^{N}\sum_{j=1}^{L_i}\mathbb{I}(s^i_j\in \{{\rm EA,IA}\})}{\sum_{i=1}^{N}\sum_{j=1}^{L_i}\mathbb{I}(s^i_j \in \{{\rm EA,IA,AA}\})}
\end{equation}

\textbf{DISTINCT} We use the DISTINCT-2 score~\cite{li-etal-2016-diversity} to calculate the proportion of repetitive parts in a dialogue. This serves as an important basis for assessing whether the conversation is progressing normally by measuring the uniqueness and variability of the dialogue exchanges.

\subsubsection{Doctor LLMs Human Evaluation Metrics}
Based on the patient simulation validation, we conducted a comparative evaluation through multi-round diagnostic dialogues, which are generated by interaction between the doctor LLMs and the SAPS. In line with the criteria proposed by the UK General Medical Council Patient Questionnaire~(GMCPQ) and principles published in reviews of the consensus for best practices for patient-centered communication (PCCBP) in medical interviews~\citep{king2013best}, we established five metrics each from the perspectives of the doctor and patient to score the consultation dialogues. 

From a doctor's perspective, Inquiry measures the capacity to gather information, such as which model more effectively collects key patient information, including chief complaints and illness history. Logic assesses which model has a more logical and non-repetitive questioning approach. Diagnosis determines which model makes accurate diagnoses with adequate information and provides appropriate advice when information is scarce. Patient evaluates which model better demonstrates empathy, respect, and support for the patient's emotional and psychological needs. Total decide which model excels overall, considering their information gathering, inquiry logic, diagnostic accuracy, and humanistic care.

From a patient's perspective, Effective measures which model provided more beneficial advice or diagnosis in general terms. Clear assesses which model communicated more clearly and was easier to understand, particularly in explaining any medical or technical terms. Understand determines which model showed greater consideration of the patient’s preferences and engagement with the patient’s ideas or concerns. Empathy evaluates which model demonstrated more empathy and a better response to the patient’s emotional state and thoughts and Total evaluates which model appeared more credible, reliable, and professional overall.

\subsection{Related Works}
In the realm of medical language model development, significant strides have been made by prior works such as HuatuoGPT~\cite{zhang2023huatuogpt} and Disc-medllm~\cite{bao2023disc}. HuatuoGPT employs a method that uses multi-turn consultation conversations generated by ChatGPT-like models for training data, whereas Disc-medllm simplifies the dialogue process into three distinct phases: information inquiry, preliminary diagnosis, and treatment suggestion. Despite the contributions of these works, they do not explore the validation of the dynamic interaction evaluation method nor establish the Automatic Interactive Evaluation (AIE) framework as a standard in medical LLMs. This validation and standardization form the crux of our research. Furthermore, while our research shares motivations with studies referenced in~\cite{johri2024craftmd,liao2023automatic}, it distinctively introduces several advancements. Notably, our paper proposes a state-aware patient simulator (SAPS) within the AIE framework, a novel development that enhances the fidelity and utility of these simulations. We also define a comprehensive action space for the doctor LLM, which is effectively utilized by the SAPS during evaluations and is crucial for computing innovative metrics. Our research extends beyond theoretical proposals by conducting rigorous patient simulator test experiments to ascertain the functionality and performance of the SAPS. Additionally, we undertake human evaluation studies to demonstrate the alignment between our proposed automatic evaluation metrics and traditional human assessment methods. This holistic approach confirms the effectiveness of our framework and underscores the significant enhancements our work offers over similar studies.

\subsection{Ethic Consideration}
This study focuses on the automatic interactive evaluation of medical consultations for Large Language Models (LLMs), which have complex ethical landscapes intersecting artificial intelligence and healthcare. Therefore, We discuss several key ethical considerations to ensure responsible research practices. 

\paragraph{Potential Risks}
The development and evaluation of LLMs in healthcare inevitably raise concerns about algorithmic biases that may perpetuate disparities in medical outcomes. We sample cases from the case library in an unbiased manner and also conduct experiments on multiple public medical examination questions to increase the diversity of cases, which can alleviate bias in the experiment results. In recognition of the critical need for transparency in healthcare interventions, we strived for explainability in the AI models' decisions and diagnoses. We propose a new approach that involves using simulated patients to model and record the behavior of doctor LLMs during the consultation process. This facilitates our study of the decision-making changes of large language models in medical consultation dialogues. Our research involves the simulation of patient-doctor interactions, raising ethical questions about patient autonomy and informed consent. While direct patient interaction was not part of the study, we emphasize the importance of obtaining informed consent in future applications that involve real patient interactions, ensuring patients are fully aware of and comfortable with the use of AI in their care. Recognizing the potential risks associated with AI in healthcare, this study upholds the principle that AI should complement, not replace, medical professionals. The LLMs are designed to support, not substitute, the diagnostic reasoning and decision-making of healthcare providers. Professional oversight is crucial in interpreting and acting upon AI-generated insights, ensuring the safe and effective use of AI technologies in patient care.

\paragraph{Data Ethics and Privacy Compliance}
Given the medical context of our research, we ensure the privacy and confidentiality of patient data used for evaluating the LLMs. We have taken stringent measures to ensure the privacy and confidentiality of this information. All personal identifiers have been removed to maintain anonymity, ensuring no individual can be recognized from the data used. During the data collection, patients signed informed consent forms and were fully aware of the data usage methods described in this paper. Additionally, the usage of this data has been reviewed and approved by the corresponding hospital ethics committees. The specific approval numbers will be provided after the end of the review. This ensures that the data usage in this paper fully complies with ethical standards and privacy protection regulations.

\subsection{Data Availability}
MedQA data is available at \url{https://github.com/jind11/MedQA}. MedMCQA data is available at \url{https://github.com/MedMCQA/MedMCQA}. MMLU data is available at \url{https://github.com/hendrycks/test}. SelfExam data is available at \url{https://journals.plos.org/digitalhealth/article?id=10.1371/journal.pdig.0000198#references}. All the experiment's preprocessed data are available in the code repository.

\subsection{Model Availability}
For the weak models, we use the \texttt{ChatGLM3-6B} available at {\url{https://github.com/THUDM/ChatGLM3}} for ChatGLM3-6b, \texttt{Baichuan-13B-Chat} available at {\url{https://github.com/baichuan-inc/Baichuan-13B}} for BaiChuan-13b, and \texttt{InternLM2-Chat-20B} available at {\url{https://github.com/InternLM/InternLM}} for InternLM-20b.

\subsection{Code Availability}
The final annotation guidelines and all synthetic datasets used in this study are
available at \url{https://github.com/BlueZeros/Automatic_Interactive_Evaluation}.

\bibliography{colm2024_conference}
\bibliographystyle{colm2024_conference}

% \appendix
% \section{Appendix}

\end{document}